\documentclass[floatfix,aps,pra,twocolumn,superscriptaddress,showpacs,a4paper,english]{revtex4}
\usepackage{graphicx,subfigure,epsfig,verbatim,psfrag,amsmath,amssymb,color,float}
\usepackage{indentfirst}
\usepackage{textcomp}
\usepackage{latexsym}
\usepackage{amssymb}
\usepackage{amsmath}
\usepackage{lipsum}
\usepackage{soul}
\usepackage{xcolor}
\usepackage{color}
\usepackage{epstopdf}
\usepackage{placeins}
\usepackage{booktabs}
\usepackage{algorithm}
\usepackage{algpseudocode}  
\usepackage{indentfirst}
\usepackage{subfigure}
\usepackage{epstopdf}
\DeclareGraphicsExtensions{.png,.eps,.pdf}
\graphicspath{{picture/}}

\makeatletter
\newenvironment{breakablealgorithm}
  {
   \begin{center}
     \refstepcounter{algorithm}
     \hrule height.8pt depth0pt \kern2pt
     \renewcommand{\caption}[2][\relax]{
       {\raggedright\textbf{\ALG@name~\thealgorithm} ##2\par}
       \ifx\relax##1\relax 
         \addcontentsline{loa}{algorithm}{\protect\numberline{\thealgorithm}##2}
       \else 
         \addcontentsline{loa}{algorithm}{\protect\numberline{\thealgorithm}##1}
       \fi
       \kern2pt\hrule\kern2pt
     }
  }{
     \kern2pt\hrule\relax
   \end{center}
  }
\makeatother

\begin{document}
	\title{Generative Tensor Network Classification Model for Supervised Machine Learning}
	
	\author{Zheng-Zhi Sun}
	\affiliation{School of Physical Sciences, University of Chinese Academy of Sciences, P. O. Box 4588, Beijing 100049, China}

	\author{Cheng Peng}
	\affiliation{School of Physical Sciences, University of Chinese Academy of Sciences, P. O. Box 4588, Beijing 100049, China}
	
	\author{Ding Liu}
	\affiliation{School of Computer Science and Technology, Tianjin Polytechnic University, Tianjin, 300387,China}
	
	\author{Shi-Ju Ran}
	\email[Corresponding author. Email: ] {sjran@cnu.edu.cn}
	\affiliation{Department of Physics, Capital Normal University, Beijing 100048, China}

	\author{Gang Su}
   	 \email[Corresponding author. Email: ] {gsu@ucas.ac.cn}
	\affiliation{Kavli Institute for Theoretical Sciences, and CAS Center for Excellence in Topological Quantum Computation, University of Chinese Academy of Sciences, Beijing 100190, China}
	\begin{abstract}
		Tensor network (TN) has recently triggered extensive interests in developing machine-learning models in quantum many-body Hilbert space. Here we purpose a generative TN classification (GTNC) approach for supervised learning. The strategy is to train the generative TN for each class of the samples to construct the classifiers. The classification is implemented by comparing the distance in the many-body Hilbert space. The numerical experiments by GTNC show impressive performance on the MNIST and Fashion-MNIST dataset. The testing accuracy is competitive to the state-of-the-art convolutional neural network while higher than the naive Bayes classifier (a generative classifier) and support vector machine. Moreover, GTNC is more efficient than the existing TN models that are in general discriminative. By investigating the distances in the many-body Hilbert space, we find that (a) the samples are naturally clustering in such a space; and (b) bounding the bond dimensions of the TN's to finite values corresponds to removing redundant information in the image recognition. These two characters make GTNC an adaptive and universal model of excellent performance.
		
	\end{abstract}
	\maketitle
	
	\section{Introduction}
    Machine learning incorporating with the principles of quantum mechanics forms a novel interdisciplinary field known as quantum machine learning \cite{biamonte2017quantum}. Among many sub-directions, machine learning in \textit{quantum space} is currently under hot debate. The quantum space, also called Hilbert space or quantum-enhanced feature space, is where the quantum states and operators live. The quantum classifiers defined in such a space have been proposed, which are expected to work on quantum hardwares such as superconducting processors \cite{havlicek2018supervised, PhysRevLett.122.040504}.
       
   In recent years, booming progresses have been made by combining quantum physics and machine learning through tensor network (TN) \cite{MAL-059, MAL-067, glasser2018supervised}. TN is a powerful tool that originates from quantum many-body physics and quantum information sciences; it can be applied to efficiently deal with the states and operators defined in many-body Hilbert space whose dimension increases exponentially with the number of sites (or physical particles) \cite{verstraete2008matrix, ORUS2014117, ran2017review, Evenbly2011, bridgeman2017hand, SCHOLLWOCK201196,Cirac_2009}. As a novel extension, TN is considered as a universal model for supervised and unsupervised learning \cite{NIPS2016_6211, liu2017machine, liu2018learning, PhysRevX.8.031012,e20080583,pestun2017tensor,cheng2019tree,8406391,PhysRevE.98.042114}. Its applications on, e.g., image recognition, already exhibit competitive performance to the conventional models such as neural networks. 
   
    With the underlying connections between TN and quantum circuits \cite{kak1995quantum, PhysRevLett.122.040504, benedetti2018generative, mitarai2018quantum, PhysRevLett.113.130503, lloyd2013quantum, PhysRevA.98.062324, zeng2018learning, mcclean2016theory, farhi2014quantum, farhi2018classification}, TN sheds new lights on quantum computation of machine learning tasks \cite{schuld2015introduction, Arunachalam:2017:GCS:3106700.3106710, PhysRevLett.109.050505}. For instance, a training algorithm  \cite{liu2017machine} inspired by the multiscale entanglement renormalization ansatz \cite{PhysRevLett.100.240603} allows using unitary gates or isometries to construct the TN for machine learning. Most recently, Huggins \textit{et al.} proposed that the quantum circuit corresponding to a TN can be easily designed by one- and two-qubit unitary gates \cite{huggins2018towards}. The appealing perspective of TN in quantum machine learning urges us to understand deeply the underlying characters of the TN machine learning, and to develop novel TN approaches of higher performance. However, some fundamental questions are still untouched. Among others, it is elusive about possible advantages of TN machine learning in quantum many-body space, compared with the models (e.g., neural network) that learn the data in the original multiple-scalar space.
   
\begin{figure}
		\includegraphics[width=0.9\linewidth]{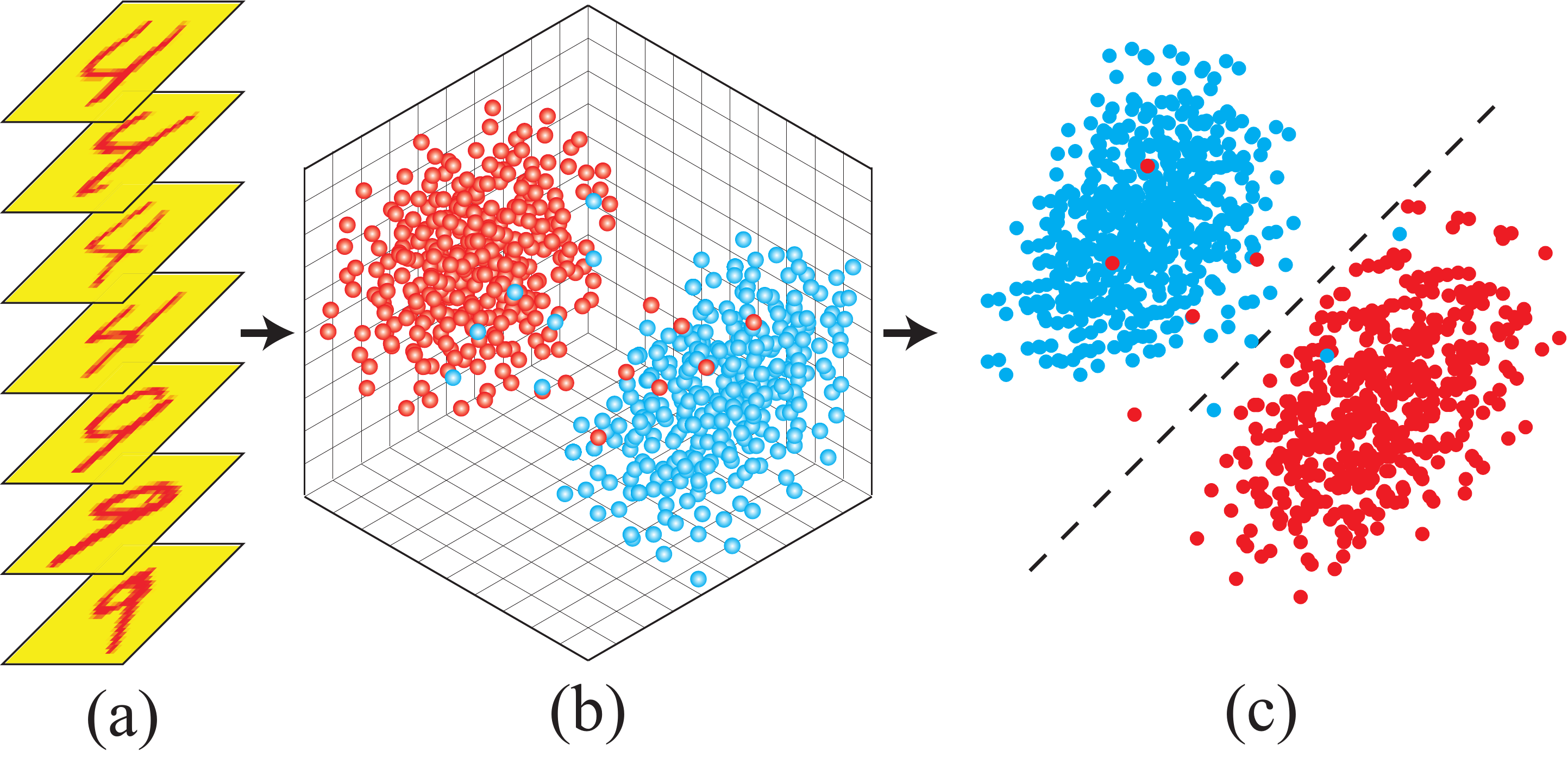}
		\caption{\label{fig-instance} (a) The images to be classified (digits `4' and `9', for example). (b) A sketch of the distribution of the samples after mapping to the many-body Hilbert space $\mathcal{H}$ by the given feature map. Since the exponentially large space cannot be shown in a figure, we use the three-dimensional space instead just for illustration. Note the blue and red crosses stand for digits '4' and '9', respectively. (c) The distribution of images after mapping to the $(\ln f_4- \ln f_9)$ space, with $\ln f_k$ the logarithmic fidelity between the sample and the generative TN of the $k$-th class. The dash line gives the boundary for classification, on which we have $f_4 = f_9$.}
	\end{figure}

	In this work, we propose a generative TN classification (GTNC) model for supervised learning (Fig.\ref{fig-instance}) in many-body Hilbert space (denoted as $\mathcal{H}$). The GTNC is formed by several generative TN's; each generative TN is a quantum state defined in $\mathcal{H}$ and is trained as the generative model for the corresponding class of images \cite{PhysRevX.8.031012}.  For a given sample, the classification is done by finding the generative TN with the smallest Euclidean distance (fidelity) to this sample in $\mathcal{H}$. In other words,  the classification is given by the boundary, from which the Euclidean distances to the generative TN's are equal. With the MNIST \cite{6296535} and fashion-MNIST \cite{xiao2017fashion} datasets, the GTNC shows remarkable efficiency and accuracy by comparing with several existing methods including the discriminative TN machine learning method \cite{NIPS2016_6211}, supportive vector machines (SVM's) \cite{Cortes1995}, and naive Bayes classifiers \cite{rish2001empirical}.
	
	Two key advantageous characters of GTNC are discussed. Firstly by computing the Euclidean distances (i.e., fidelity) among the samples and the generative TN's, we show that the samples mapped to the many-body Hilbert space are naturally clustering. It implies that the classification can be efficiently and accurately done in such a space. The clustering should be an advantage from the space $\mathcal{H}$. Though the idea of mapping to such a space of a much higher dimension is analogous to SVM, better accuracy is achieved with GTNC. Secondly by comparing with a lazy-learning baseline model, we show that bounding the bond dimensions of the TN's to finite values corresponds to removing redundant information in the image recognition. The relation to the quantum entanglement can be addressed.
	
	\section{Generative tensor network classification algorithm for supervised learning}
	
	The training of GTNC is to obtain the generative TN $\mathbf{\Psi}^{(c)}$ ($c= 1, \cdots, K$ with $K$ the total number of classes in the classification task) for each of the classes. We use the algorithm proposed by Han \textit{et al.} \cite{PhysRevX.8.031012}. To begin with, one builds a one-to-one map called feature map \cite{NIPS2016_6211,novikov2016exponential}, which maps the images to a vector space known as many-body Hilbert space (denoted as $\mathcal{H}$) in quantum physics. For example, the feature map that transforms the $i$-th pixel $x_i$ (normalized so that $0 \le {x_i} \le 1$) to a two-component vector can be written as
	\begin{equation}
	\label{eq1}
	\mathbf{s}_i = \left[ {\cos \left( {{\pi  \over 2}{x_i}} \right),\sin \left( {{\pi  \over 2}{x_i}} \right)} \right]^T.
	\end{equation}
	In this way, one image that consists of $L$ pixels is mapped to the direct product of $L$ vectors as $\mathbf{v} = \prod_{\otimes i} {{\mathbf{s}_i}} $. Physically, $\mathbf{v}$ can be regarded as the product state of $L$ qubits. Each qubit has two components, equivalent to a spin-$1/2$. Note that it is possible to generalize the feature map to be $d$-component with $d>2$. Then one image is mapped to a vector defined in the $d^L$-dimensional vector space. 
	
	TN is utilized as the generative model of each class \cite{verstraete2008matrix, cirac2009renormalization, SCHOLLWOCK201196, ORUS2014117}. In fact, the generative models are quantum states of $L$ bodies defined in $\mathcal{H}$, which capture the joint probability distributions of the corresponding sets. In quantum many-body physics, TN has been shown as an efficient and power tool to deal with quantum many-body states, where the computational complexity can be reduced from exponentially-hard to polynomial-hard. In Ref. \cite{PhysRevX.8.031012}, the generative TN is trained with a gradient algorithm that minimizes the cost function of Kullback–Leibler divergence \cite{10.2307/2236703}. 
	
	After training the generative TN's $\{\mathbf{\Psi}^{(c)}\}$, a given sample can be classified by comparing the Euclidean distances in $\mathcal{H}$ between this sample and $\{\mathbf{\Psi}^{(c)}\}$. We choose the fidelity $f_c$ to measure the distance, which is defined as
	\begin{equation}\label{eq-fidelity}
	f_c =  |\mathbf{v}^{\dagger} \mathbf{\Psi}^{(c)}|,
	\end{equation}
	with $\mathbf{v}$ the sample after the feature map. Note that in quantum information, fidelity is a measurement of distance between two quantum states. The classification is indicated by finding the largest fidelity, i.e., $\arg\max_c f_c$. One can find the pseudo code of GTNC in Sec.\ref{sec.TNmethod}.
    
	\section{Experiments}
	
	\subsection{GTNC: an adaptive generative classification model}
	
	\begin{figure}
	\includegraphics[width=1\linewidth]{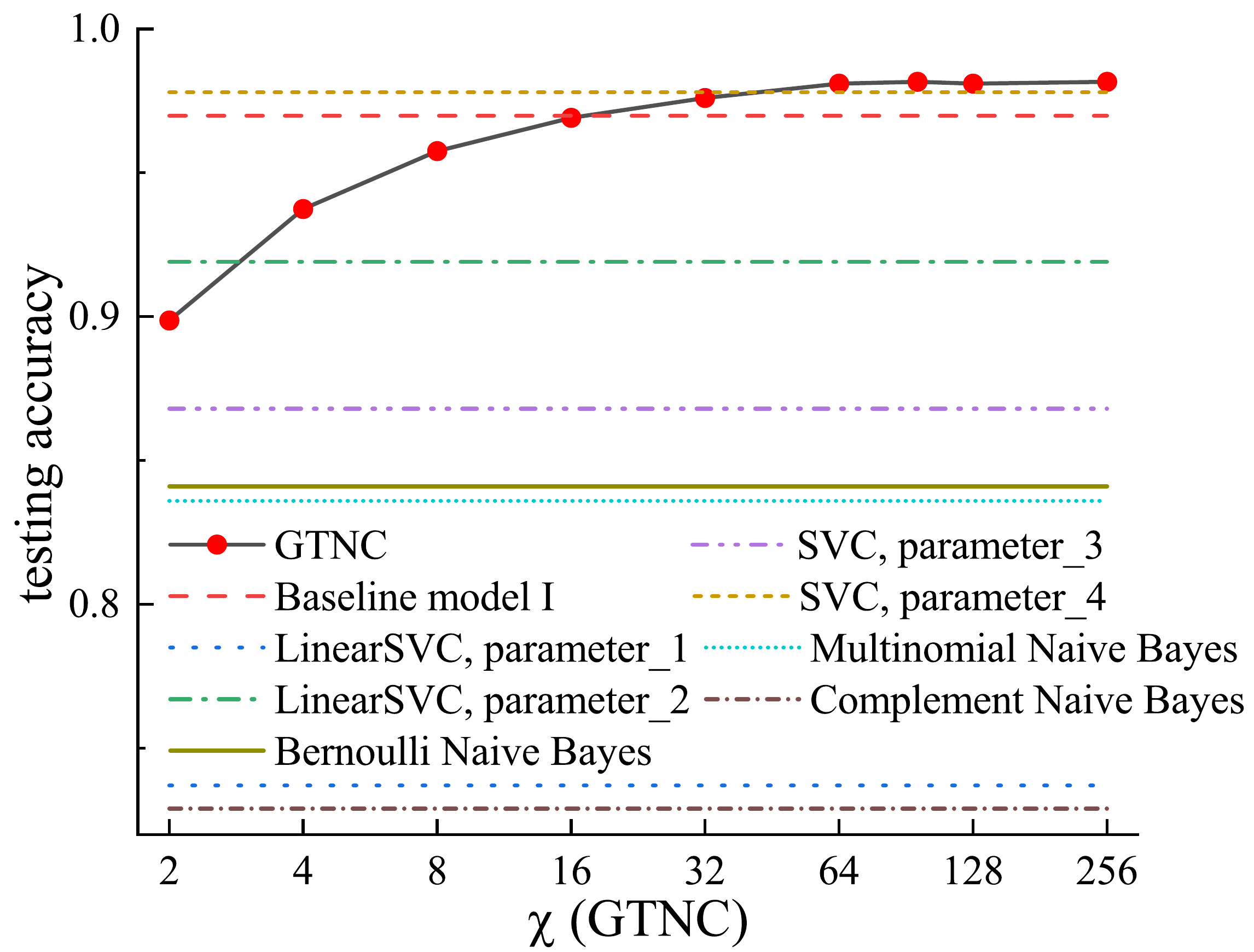}
	\caption{\label{fig-accuracy-svm-gc}The testing accuracy of MNIST dataset using GTNC, baseline model \uppercase\expandafter{\romannumeral1}, naive Bayes classifiers and SVM's. $\chi $ is the parameter that determines the complexity of GTNC. The hyper-parameters in SVM's are taken as the following: (1) loss=hinge, C=100, multi-class=crammer-singer, penalty=12; (2) loss=hinge, C=1, multi-class=crammer-singer, penalty=12; (3) C=100, kernel=sigmoid; (4) C=100, kernel=poly.}
	\end{figure}

	On the MNIST dataset, GTNC is compared with other well-established methods (Fig.\ref{fig-accuracy-svm-gc}), i.e., classical generative classifiers (naive Bayes classifiers), high-dimensional classifiers (SVM's) and a baseline model which is a lazy-learning model using feature map without TN (see Eq.\ref{eq-lazy}). For GTNC, different bond dimensions $\chi$ are taken, which controls the number of variational parameters (see Sec.\ref{sec.TNmethod} for details). We also testify on the fashion-MNIST. The 10-class testing accuracy of GTNC reaches around $88.2\%$, while for the SVM's it varies from  $48.4\% $ to $89.7 \% $ depending on the parameters \cite{xiao2017fashion}. For the naive Bayes classifiers, the testing accuracy is no higher than $70.6\%$ for the fashion-MNIST dataset.
	
	The experiments given by Fig. \ref{fig-accuracy-svm-gc} reveal several advantages of GTNC. One is that the accuracy of GTNC significantly surpasses the naive Bayes classifiers. Note that it is usual to use the discriminative models to do classification, such as convolutional neural networks (CNN's). The state-of-the-art accuracy is $99.77\%$ \cite{Ciresan2012MulticolumnDN} for MNIST and $92.54\%$ \cite{8313740} fashion-MNIST, respectively. It is true that the accuracy of GTNC is competitive but still lower than the best discriminative models. Nevertheless, The previous knowledge is necessary for these discriminative models (such as architecture and other hyper-parameters that can largely affect the results) to reach the best accuracy. For GTNC in contrast, we use the same architecture of the TN (a 1D TN that is the same as Ref. \cite{PhysRevX.8.031012}) and the same hyper-parameters (such as the feature map) for different datasets. We are optimistic to improve further the accuracy of GTNC by optimizing  the architecture and hyper-parameters, which however is beyond the scope of the current work.
	
	Secondly, GTNC possesses striking resemblance as well as essential differences compared with the SVM's. The main idea of SVM is to map the samples to a much higher dimensional space, where it becomes relatively easy to find the boundary for classification. For GTNC, the feature map is to map the samples to the many-body Hilbert space that is exponentially large. By training the generative TN's in such a space, the boundary for classification is found by computing the fidelity.
	
	It is the underlying difference that makes GTNC superior than SVM. The first difference concerns the kernel function and the space. In most cases of SVM, the mapping method is implicit and determined by a positive-defined matrix which is the distance in the higher dimensional space. This distance matrix is calculated by a certain kernel function like radial basis function kernel with origin data. Mercer's condition can guarantee that the kernel function will correspond to a higher dimensional space \cite{doi:10.1098/rsta.1909.0016}. The mapping method is implicit, thus it becomes extremely challenging to analyse how to improve the performance of SVM. The results strongly depend on the space to where the data are mapped, and the hyper-parameters such as the soft margin \cite{Cortes1995}. There is no general theories of finding the best parameters of SVM \cite{leslie2001spectrum}, which hinders the applications of SVM to new challenging problems.
	
	In comparison, GTNC is more universal and less parameter-dependent. The kernel function in GTNC is determined by the feature map and can be explicitly written, which satisfies Mercer's condition. For different datasets, we use the same feature map [see Eq. (\ref{eq1})] to transform the data to the higher-dimensional space. It is possible to optimize the  kernel function of GTNC to further improve the performance.
	
	The general strategies of GTNC and SVM are also different.  The GTNC is formed by several generative models, each of which learns the joint probability distribution of one class of samples. The classification is determined by taking the generative models as the references. Such a strategy works well due to the clustering of the samples in $\mathcal{H}$, giving higher accuracy than the naive Bayes classifiers. It is also avoided to input the samples of all classes at the same time to train the classifier(s), which leads to higher efficiency compared with the discriminative algorithms. For SVM, it is to find the classification boundary in the higher-dimensional space. This might also be one reason that the results of SVM largely depends on the chosen space and the hyper-parameters.

	We shall note when we build a SVM model with the kernel function from the feature map, the testing accuracy is extremely poor (no more than $30.0\%$). This suggests that the kernel from the feature map works with the algorithms of SVM much worse than the generative TN algorithm \cite{PhysRevX.8.031012}.

	Thirdly, the accuracy of GTNC also surpasses the baseline model \uppercase\expandafter{\romannumeral1} with a moderate bond dimension ($1 \ll \chi \ll d^L$ with $d^L$ the dimension of $\mathcal{H}$). The baseline model \uppercase\expandafter{\romannumeral1} is a lazy-learning version of GTNC; the generative model $\tilde{\mathbf{\Psi}}$ of the $c$-th class can be defined as
	\begin{equation}\label{eq-lazy}
	\tilde{\mathbf{\Psi}}^{(c)} = \sum_{\mathbf{v} \in c} \mathbf{v} / \sqrt{N_c},
	\end{equation}	
	with $N_c$ the number of samples in the $c$-th class. It means ${\widetilde {\bf{\Psi }}^{(c)}}$ is simply the summation of all vectorized samples in the $c$-th class. 
	
	For classifying a given sample $\mathbf{v}$, we still use Eq. (\ref{eq-fidelity}) to define the fidelity as $\tilde{f}_c = |\mathbf{v}^{\dagger} \tilde{\mathbf{\Psi}}^{(c)}|$. The classification of $\mathbf{v}$ is given by $\arg\max_{c} \tilde{f}_c$. Different from GTNC, we do not need to train $\tilde{\mathbf{\Psi}}$ to classify. The fidelity can be directly calculated as $\tilde{f}_c = |\sum_{\mathbf{u} \in c} \mathbf{v}^{\dagger} \mathbf{u}| / \sqrt{N_c}$, which makes baseline model \uppercase\expandafter{\romannumeral1} a lazy-learning model.
	
	Let us consider to write $\tilde{\mathbf{\Psi}}$ in a TN form just like $\mathbf{\Psi}$ in GTNC. It is expected that the bond dimension of $\tilde{\mathbf{\Psi}}$ should be extremely large. In other words, $\mathbf{\Psi}$ in GTNC can be understood as a finite-bond-dimensional approximation of $\tilde{\mathbf{\Psi}}$. Surprisingly, the accuracy of GTNC is higher than the baseline model \uppercase\expandafter{\romannumeral1}. It implies that by taking a moderate bond dimension, some redundant information are removed and a better classification can then be made.

	The value of $\chi$ actually characterizes the capacity of quantum entanglement that the TN (state) can carry. The entanglement entropy of the TN here is the R\'enyi entropy of the dataset, which is defined as ${H_\alpha } = {1 \over {1 - \alpha }}\sum\limits_{i = 1}^n {\log \left( {p_i^\alpha } \right)} $ with  ${p_i}$ the probabilities to have the $i$-th sample and $\alpha$ a constant \cite{renyi1961measures}. The entanglement entropy corresponds to the case of $\alpha=2$. The R\'enyi entropy satisfies ${H_2}\left( \chi  \right) \le \log \left( \chi  \right)$. In other words, by reducing $\chi$, the maximum of R\'enyi entropy becomes smaller. The regularization process in GTNC (known as canonicalization \cite{PhysRevB.78.155117}) guarantees that one always discards the less entangled basis. A former work showed that the less-entangled sites (pixels) contain less-important information, which can be discarded without harming too much the accuracy \cite{liu2018learning} It means the TN machine learning can be implemented more efficiently with a much smaller number of features. In accordance, our experiments demonstrate that the important information is restored in the highly-entangled basis. It is suggested that with the same number of features, the number of variational parameters in the TN can be safely reduced by removing the less-entangled basis. This shows that the over-fitting of the TN machine learning can be avoided in a controllable manner according to the quantum entanglement. 
		
	\subsection{Natural clustering}
	
\begin{figure}
		\includegraphics[width=1\linewidth]{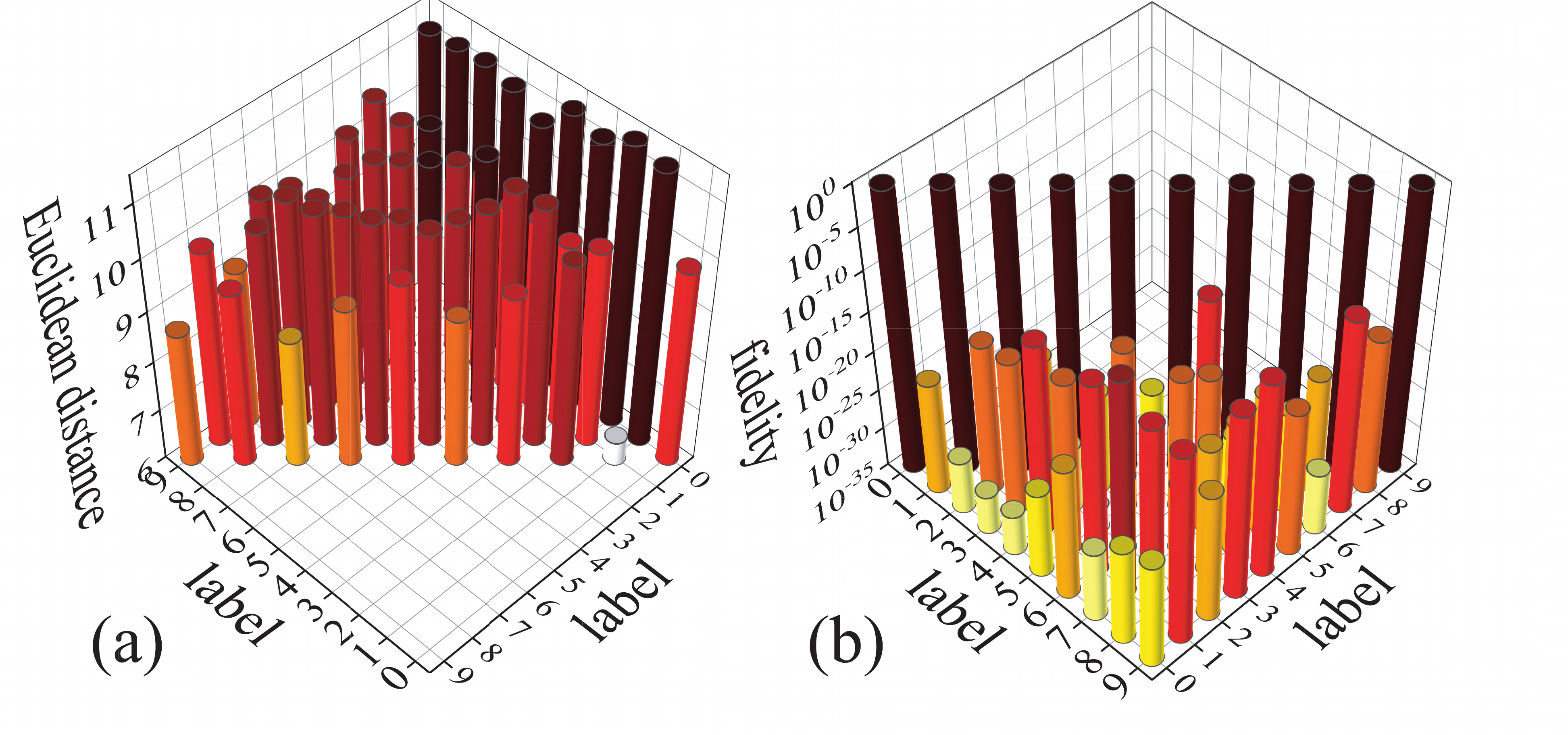}
		\caption{\label{fig-fidelity-all}(a) Euclidean Distance between samples of MNIST dataset in original space. (b) Fidelities between samples of MNIST dataset in exponentially large space.}
	\end{figure}

	To further understand the GTNC, we calculate the distances of the training samples in different spaces (Fig.\ref{fig-fidelity-all}). The Euclidean distance between the $c_1$ and $c_2$-th classes in original multi-scalar space is defined as
	\begin{equation}\label{eu-distance}
	{D_{{c_1}{c_2}}} = {1 \over {{N_{{c_1}}}{N_{{c_2}}}}}\sum\limits_{x \in {c_1}} {\sum\limits_{y \in {c_2}} {\sqrt {\sum\limits_{i = 1}^L {{{\left( {{x_i} - {y_i}} \right)}^2}} } } },
	\end{equation}
	where the pixels are normalized as $0 \leq x \leq1$. In the many-body Hilbert space $\mathcal{H}$, the fidelity is used to represent distance of two classes, which is defined as
	\begin{equation}\label{fide-distance}
   	F_{{c_1}{c_2}} = \frac{\sum_{\mathbf{u} \in c_1} \sum_{\mathbf{v} \in c_2} \mathbf{u}^{\dagger} \mathbf{v}} {\sqrt{N_{c_1} N_{c_2}}}.
	\end{equation}
	$F_{{c_1}{c_2}}$ characterizes the closeness of two classes of images.
	
	In the original space, the distances are at the same order of magnitude for the samples of the same class or of two different classes [Fig.\ref{fig-fidelity-all} (a)]. It means the distribution of the samples in this space is more or less random. In many-body Hilbert space $\mathcal{H}$, the fidelity in $\mathcal{H}$ between different classes is over $10^5$ times lower than those with the same label (the diagonal terms $F_{cc} \simeq 1$) [Fig.\ref{fig-fidelity-all} (b)]. In other words, the distance between the samples of different classes is averagely much larger than that between the samples of a same class. This means the samples of the same class are clustering in $\mathcal{H}$, which makes it much easier to classify. 
	
	The clustering is also consistent with the fair accuracy of the baseline model \uppercase\expandafter{\romannumeral1}. Let us rewrite $F_{{c_1}{c_2}}$ in terms of the generative TN's of the baseline model \uppercase\expandafter{\romannumeral1} [Eq. (\ref{eq-lazy})]. We have ${F_{{c_1}{c_2}}} = |{{\bf{\tilde \Psi }}^{({c_1})\dag }}{{\bf{\tilde \Psi }}^{({c_2})}}|$, which is the fidelity of the two generative TN's. As the samples are clustering, the distance from a given sample to the correct ${\bf{\tilde \Psi }}^{({c})}$ should be much smaller than that to a wrong one. Thus, the classification can be accurately done by comparing the distances. Note the Euclidean distance in $\mathcal{H}$ can be deducted from fidelity, satisfying $D_{c_1 c_2} = |\mathbf{\tilde{\Psi}}^{(c1)} - \mathbf{\tilde{\Psi}}^{(c2)}|^2 = F_{c_1c_1}^2 + F_{c_2c_2}^2 - 2F_{{c_1}{c_2}}$ with $F_{cc} = |\mathbf{\tilde{\Psi}}_{c}|$. We do not enforce the normalization of $\{\mathbf{\tilde{\Psi}}^{(c)}\}$, though we have  $|\mathbf{\tilde{\Psi}}_{c}| \simeq 1$, giving $D_{c_1 c_2} \simeq 2 - 2F_{{c_1}{c_2}}$.
	
	We also compare GTNC with th existing discriminative TN model (dubbed as the baseline model \uppercase\expandafter{\romannumeral2}) \cite{NIPS2016_6211}. The pseudo-code can be found in the Sec.\ref{sec.code}. Our experiments show that the efficiency of GTNC is significantly higher than the baseline model \uppercase\expandafter{\romannumeral2}. On MNIST dataset with $\chi=32$, the accuracy of GTNC converges to $97.6\%$ in about $9 \times {10^3}$ seconds of CPU time, while the accuracy of the baseline model \uppercase\expandafter{\romannumeral2} converges to $87.3\%$ in about $4 \times {10^6}$ seconds \footnote{The CPU model is Intel(R) Xeon(R) CPU E5-2630 v3 @ 2.40GHz.} . The efficiency differs due to the strategy. For GTNC, one will only input one class of images to train each of the generative TN, and the tensors converge with a small number of iterations. For the baseline model \uppercase\expandafter{\romannumeral2}, one will input the samples of all classes to train the classifier, and it needs much more iterations to converge. The complexity analysis shows that even in one iteration, the computational complexity of GTNC is much lower than baseline model \uppercase\expandafter{\romannumeral2} because the samples only need to be input into the corresponding tensor network in GTNC.

	\section{Discussion and Perspective}
	
     In this work, we propose the generative TN classification (GTNC) method, and based on it investigate several fundamental issues of the TN machine learning, i.e., the roles played by the feature map and by the bond dimensions of the TN representation. The main contributions of this work are concluded in the following.
     \begin{itemize}
     	\item GTNC is proposed as a generative model for supervised machine learning. The central idea is to individually train the generative TN's in many-body Hilbert space for samples with different labels, and to classify by comparing the distances. The performance of GTNC surpasses the existing (discriminative) TN-machine learning methods, the Naive Bayes method which are also generative classifier, and the supportive vector machine.
     	\item The role of feature map is revealed. We show that the feature map of the TN machine learning methods is to map the samples to an exponentially large vector space (called many-body Hilbert space in physics). In such a space, the samples are naturally clustering, where the classification can be easily and accurately done with the help of the generative TN's.
     	\item The relation between entanglement and machine learning is discussed, which is useful to avoid over-fitting in a controllable way. The experiments by comparing GTNC with baseline model \uppercase\expandafter{\romannumeral1} imply that the important information is restored in the highly-entangled basis of the generative TN's. By keeping a proper number of the relatively highly-entangled basis, the accuracy surpasses the baseline model \uppercase\expandafter{\romannumeral1}, where all bases are taken into consideration.
     \end{itemize}
     
     Our work contributes to answering an important question: whether there exist any advantages to solve machine learning problems in the exponentially-large many-body Hilbert space by TN than in the multiple-scalar space by the classical machine learning models. While the previous simulations of the TN machine learning algorithms have given considerable promising results, our experiments show a positive answer in a more explicit way. Such investigations will strongly motivate to develop the quantum computation of machine learning in the many-body Hilbert space, such as the machine learning schemes by quantum circuits \cite{huggins2018towards}. The benefits or ``quantum supremacy'' will be not just limited to quantum acceleration, but also to develop more universal, powerful, and well-controlled machine learning models.

     \section{Methods}
     \subsection{Tensor network machine}
     \label{sec.TNmethod}
    The functions (vectors or operators) defined in this exponentially large space $[\mathbb{V}^d]^{\otimes L}$ might have lager potential for generating and learning compared with the multi-scalar functions (such as neural network). One problem is how to handle such an exponentially large space, which is usually NP-hard by classical computers.

\begin{figure}
	\centering
	\includegraphics[width=0.9\linewidth]{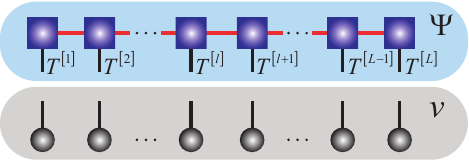}
	\caption{\label{fig-TN} Illustration of tensor network.}
	\end{figure}

    The quantum many-body physics provides us a solution called tensor network (TN) , which reduces the cost from exponential to polynomial or linear manner \cite{verstraete2008matrix,cirac2009renormalization,SCHOLLWOCK201196,ORUS2014117}. TN is an efficient representation of one (large) tensor by writing it as the contraction of several tensors. Matrix product state \cite{perez2006matrix} is one form of the TN's (Fig.\ref{fig-TN}), which has been utilized in the machine learning field \cite{PhysRevX.8.031012,NIPS2016_6211}. An MPS formed by $L$ tensors can be  written as
	\begin{equation}\label{eq-MPS}
    {\Psi _{{s_1}{s_2} \cdots {s_L}}} = \sum\limits_{{\alpha _1} \cdots {\alpha _{L - 1}}} {T_{{s_1},{\alpha _1}}^{[1]}} T_{{s_2},{\alpha _{1}}{\alpha_2}}^{[2]} \cdots T_{{s_{L-1}},{\alpha _{L-2}}{\alpha_{L-1}}}^{[L-1]} T_{{s_L},{\alpha_{L-1}}}^{[L]}.
	\end{equation}
    $T^{[l]}$ is a $(d \times \chi \times \chi)$-dimensional tensor located on the $l$-th site, the indexes $\{\alpha_l\}$ and $\{s_n\}$ ($l = 1, \cdots, L-1$) are dubbed as virtual and physical bonds, respectively. $\chi$ is called the bond dimension of the MPS, which determines the number of parameters and the upper bound of the entanglement that the MPS can capture (see for example Ref. \cite{PhysRevB.78.024410}). For simplicity, we assume all elements of the tensors are real numbers. It is easy to see that by contracting all the virtual bonds, $\mathbf{\Psi}$ is a vector in the $[\mathbb{V}^d]^{\otimes L}$ space, whose dimension increases exponentially with $L$. Thanks to the TN structure of $\mathbf{\Psi}$, the number of parameters is about $Ld\chi^2$ in the MPS, which scales only linearly with $L$.
	
	The gradient descent algorithm is used to optimize the MPS. For example, the $l$-th tensor of the MPS is updated as
    \begin{equation}\label{eq-gradient-update}
    {\mathbf{T}^{[l]}} \leftarrow \mathbf{T}^{[l]} - \eta \frac{{\partial \Gamma }}{{\partial {\mathbf{T}^{[l]}}}},
    \end{equation}
    where $\eta$ is the step of the gradient decent algorithm and $\Gamma $ is the cost function. All tensors are updated iteratively until the preset convergence is reached. One could refer to Refs. \cite{PhysRevX.8.031012,NIPS2016_6211} for more details.

	\subsection{Pseudo code of GTNC and baseline model \uppercase\expandafter{\romannumeral2}}
	\label{sec.code}
    In the GTNC, the algorithm for training the generative MPS's follows Ref. \cite{PhysRevX.8.031012}. The baseline model\uppercase\expandafter{\romannumeral1} algorithm follows Ref. \cite{NIPS2016_6211}. And the pseudo codes are shown in Al.\ref{Sun-single} and Al.\ref{miles-double}.

	\begin{breakablealgorithm}
	\caption{- GTNC}	
 	\label{Sun-single}
    	\begin{algorithmic}[1]
    	\Require
    	$\alpha$:Step size;
    	\Require
    	$\beta $:decay rate for Step size;
    	\Require
    	$t \leftarrow 0$ (Initialize time step) \par
    	$\Gamma \left( {{T^{\left[ 1 \right]}},{T^{\left[ 2 \right]}}, \cdots ,{T^{\left[ L \right]}}} \right)$: Stochastic objective function with parameters${\left\{ {{T^{\left[ l \right]}}} \right\}}$ [a]
    	\Require
    	$\left\{ {T_0^{\left[ l \right]}} \right\}$: Initial parameter vector\par
    	$\left\{ {{p_{l,0}}} \right\} \leftarrow 0$ (Initialize  moment vector)\par
    	\While{$\left\{ {T_t^{\left[ l \right]}} \right\}$not converged}
    	\State $t \leftarrow t + 1$
    	\For{$l = 1:L$}
    	\State ${g_{l,t}} \leftarrow {\left. {{{d{\Gamma _t}} \over {d{T^{\left[ l \right]}}}}} \right|_{{T^{\left[ l \right]}} = T_{t - 1}^{\left[ l \right]}}}$	
    	\State ${p_{l,t}} \leftarrow {{g_{l,t}^2} \over {{{\left| {T_{t - 1}^{\left[ l \right]}} \right|}^2}}}$
    	\State $T_t^{\left[ l \right]} \leftarrow T_{t - 1}^{\left[ l \right]} - \alpha  \cdot {{{g_{l,t}}} \over {\sqrt {{p_{l,t}}} }}$
    	\If {$l < L$}
    	\State $Q,R \leftarrow qr\left( {T_t^{\left[ l \right]}} \right)$($Q$ is an unitary matrix matrixes. $R$ is an upper triangular matrix. And they satisfy $QR = T_t^{\left[ l \right]}$)
    	\State $T_t^{\left[ l \right]} \leftarrow Q,T_t^{\left[ {l + 1} \right]} \leftarrow RT_{t - 1}^{\left[ {l + 1} \right]}$
    	\EndIf
    	\EndFor
    	\If{$\Gamma \left( {\left\{ {T_t^{\left[ l \right]}} \right\}} \right) > \Gamma \left( {\left\{ {T_{t - 1}^{\left[ l \right]}} \right\}} \right)$}
    	\State $\alpha  \leftarrow {\alpha  \mathord{\left/{\vphantom {\alpha  \beta }} \right.\kern-\nulldelimiterspace} \beta }$
    	\EndIf
    	\EndWhile
    	\State \Return $\left\{ {T_t^{\left[ l \right]}} \right\}$
    	\end{algorithmic}
    	\end{breakablealgorithm}

	\begin{breakablealgorithm}
    	\caption{- baseline model \uppercase\expandafter{\romannumeral2}}
    	\label{miles-double}
    	\begin{algorithmic}[1]
    	\Require
    	$\alpha$:Step size;
    	\Require
    	$\beta $:decay rate for Step size;
    	\Require
    	$t \leftarrow 0$ (Initialize time step) \par
    	$\Gamma \left( {{T^{\left[ 1 \right]}},{T^{\left[ 2 \right]}}, \cdots ,{T^{\left[ L \right]}}} \right)$: Stochastic objective function with parameters${\left\{ {{T^{\left[ l \right]}}} \right\}}$ [b]
    	\Require
    	$\left\{ {T_0^{\left[ l \right]}} \right\}$: Initial parameter vector\par
    	$\left\{ {{p_{l,0}}} \right\} \leftarrow 0$ (Initialize  moment vector)\par
    	\While{$\left\{ {T_t^{\left[ l \right]}} \right\}$not converged}
    	\State $t \leftarrow t + 1$
    	\For{$l = 1:L-1$}
    	\State ${g_{l,t}} \leftarrow {\left. {{{d{\Gamma _t}} \over {d{T^{\left[ {l,l + 1} \right]}}}}} \right|_{{T^{\left[ {l,l + 1} \right]}} = T_{t - 1}^{\left[ {l,l + 1} \right]}}}$
    	\State ${p_{l,t}} \leftarrow {{g_{l,t}^2} \over {{{\left| {T_{t - 1}^{\left[ {l,l + 1} \right]}} \right|}^2}}}$
    	\State $T_t^{\left[ {l,l + 1} \right]} \leftarrow T_{t - 1}^{\left[ {l,l + 1} \right]} - \alpha  \cdot {{{g_{l,t}}} \over {\sqrt {{p_{l,t}}} }}$
    	\State $U,L,V \leftarrow svd \left( {\;T_t^{\left[ {l,l + 1} \right]}} \right)$($U$ and $V$ are unitary matrixes. $L$ is diagonal matrix. And they satisfy $UL{V^\dag } = T_t^{\left[ {l,l + 1} \right]}$)
    	\State $T_t^{\left[ l \right]} \leftarrow U,T_t^{\left[ {l + 1} \right]} \leftarrow L{V^\dag }$ [The process of moving label is shown in Fig. \ref{fig-move-label}]
    	\EndFor
    	\If{$\Gamma \left( {\left\{ {T_t^{\left[ n \right]}} \right\}} \right) > \Gamma \left( {\left\{ {T_{t - 1}^{\left[ n \right]}} \right\}} \right)$}
    	\State $\alpha  \leftarrow {\alpha  \mathord{\left/{\vphantom {\alpha  \beta }} \right.\kern-\nulldelimiterspace} \beta }$
    	\EndIf
    	\EndWhile
    	\State \Return $\left\{ {T_t^{\left[ l \right]}} \right\}$
    	\end{algorithmic}
    	\end{breakablealgorithm}
    	
    	Note: [a] The cost function is chosen as follow
    	\begin{equation}
    	\label{eq-cross}
    	\Gamma  =  - {1 \over J}\sum\limits_j {\ln {{{P_j}} \over Z}}  - \ln \left( J \right).
    	\end{equation}

    	Note: [b] The cost function is chosen as follow
	\begin{equation}
	\label{eq-quadratic-cf}
	\Gamma  = \sum\limits_j {{{\left| {{{\tilde L}_j} - L_j^c} \right|}^2}} .
	\end{equation}
	
    	\begin{figure}
    	\centering
    	\includegraphics[width=0.9\linewidth]{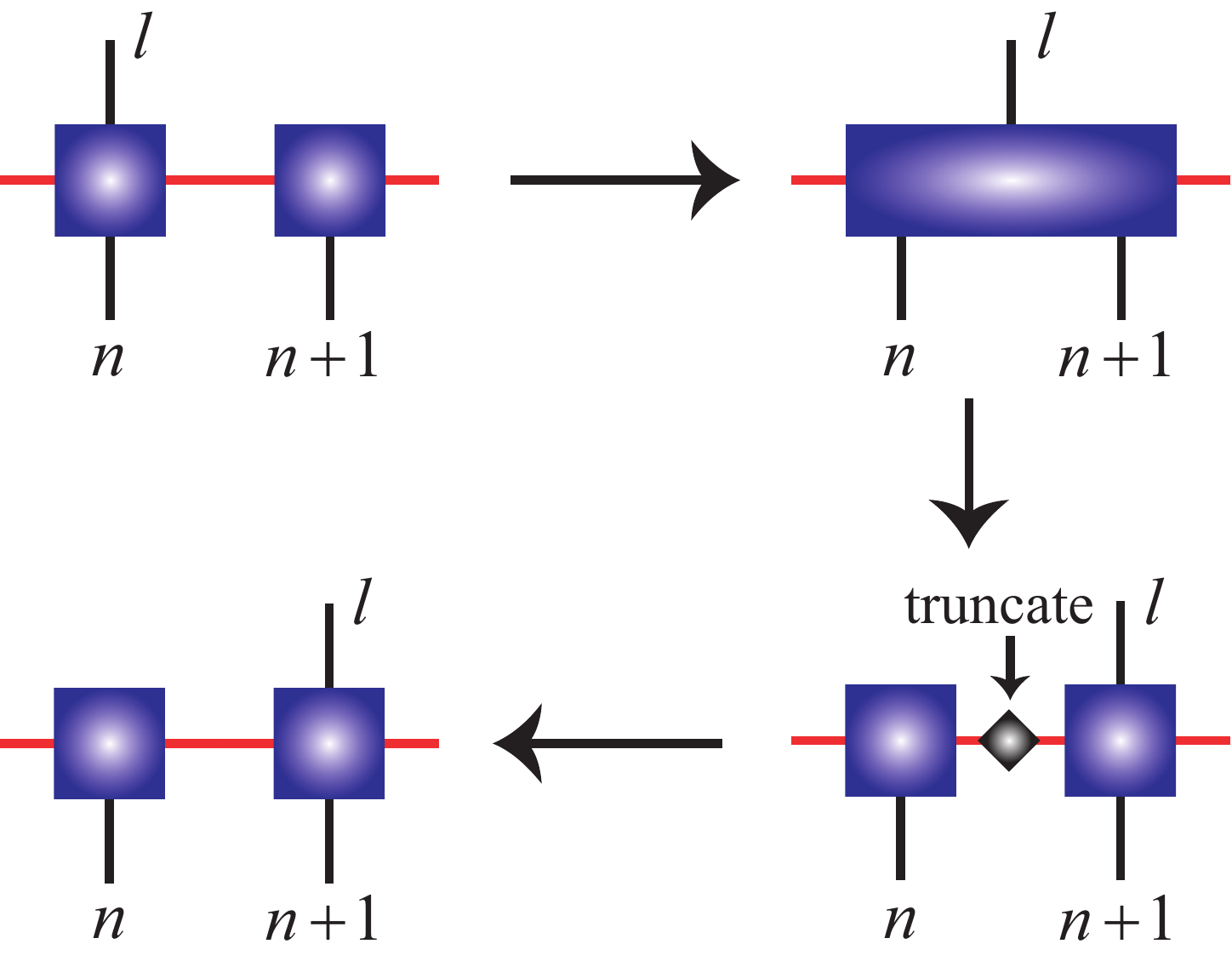}
    	\caption{\label{fig-move-label}Illustration of moving label index.}
    	\end{figure}

     \section*{Acknowledgments}

    This work was supported in part by the National Natural Science Foundation of China (11834014 and 11474279), the National Key R$\&$D Program of China (2018YFA0305800), and the Strategic Priority Research Program of the Chinese Academy of Sciences (XDB28000000). S.J.R. is supported by Beijing Natural Science Foundation (1192005  and Z180013) and Foundation of Beijing Education Committees under Grants No. KZ201810028043.


\begin{thebibliography}{52}
\expandafter\ifx\csname natexlab\endcsname\relax\def\natexlab#1{#1}\fi
\expandafter\ifx\csname bibnamefont\endcsname\relax
  \def\bibnamefont#1{#1}\fi
\expandafter\ifx\csname bibfnamefont\endcsname\relax
  \def\bibfnamefont#1{#1}\fi
\expandafter\ifx\csname citenamefont\endcsname\relax
  \def\citenamefont#1{#1}\fi
\expandafter\ifx\csname url\endcsname\relax
  \def\url#1{\texttt{#1}}\fi
\expandafter\ifx\csname urlprefix\endcsname\relax\def\urlprefix{URL }\fi
\providecommand{\bibinfo}[2]{#2}
\providecommand{\eprint}[2][]{\url{#2}}

\bibitem[{\citenamefont{Biamonte et~al.}(2017)\citenamefont{Biamonte, Wittek,
  Pancotti, Rebentrost, Wiebe, and Lloyd}}]{biamonte2017quantum}
\bibinfo{author}{\bibfnamefont{J.}~\bibnamefont{Biamonte}},
  \bibinfo{author}{\bibfnamefont{P.}~\bibnamefont{Wittek}},
  \bibinfo{author}{\bibfnamefont{N.}~\bibnamefont{Pancotti}},
  \bibinfo{author}{\bibfnamefont{P.}~\bibnamefont{Rebentrost}},
  \bibinfo{author}{\bibfnamefont{N.}~\bibnamefont{Wiebe}}, \bibnamefont{and}
  \bibinfo{author}{\bibfnamefont{S.}~\bibnamefont{Lloyd}},
  \bibinfo{journal}{Nature} \textbf{\bibinfo{volume}{549}},
  \bibinfo{pages}{195} (\bibinfo{year}{2017}).

\bibitem[{\citenamefont{Havlicek et~al.}(2019)\citenamefont{Havlicek,
  C{\'o}rcoles, Temme, Harrow, Chow, and Gambetta}}]{havlicek2018supervised}
\bibinfo{author}{\bibfnamefont{V.}~\bibnamefont{Havlicek}},
  \bibinfo{author}{\bibfnamefont{A.~D.} \bibnamefont{C{\'o}rcoles}},
  \bibinfo{author}{\bibfnamefont{K.}~\bibnamefont{Temme}},
  \bibinfo{author}{\bibfnamefont{A.~W.} \bibnamefont{Harrow}},
  \bibinfo{author}{\bibfnamefont{J.~M.} \bibnamefont{Chow}}, \bibnamefont{and}
  \bibinfo{author}{\bibfnamefont{J.~M.} \bibnamefont{Gambetta}},
  \bibinfo{journal}{Nature} \textbf{\bibinfo{volume}{567}},
  \bibinfo{pages}{209} (\bibinfo{year}{2019}).

\bibitem[{\citenamefont{Schuld and Killoran}(2019)}]{PhysRevLett.122.040504}
\bibinfo{author}{\bibfnamefont{M.}~\bibnamefont{Schuld}} \bibnamefont{and}
  \bibinfo{author}{\bibfnamefont{N.}~\bibnamefont{Killoran}},
  \bibinfo{journal}{Phys. Rev. Lett.} \textbf{\bibinfo{volume}{122}},
  \bibinfo{pages}{040504} (\bibinfo{year}{2019}).

\bibitem[{\citenamefont{Cichocki et~al.}(2016)\citenamefont{Cichocki, Lee,
  Oseledets, Phan, Zhao, and Mandic}}]{MAL-059}
\bibinfo{author}{\bibfnamefont{A.}~\bibnamefont{Cichocki}},
  \bibinfo{author}{\bibfnamefont{N.}~\bibnamefont{Lee}},
  \bibinfo{author}{\bibfnamefont{I.}~\bibnamefont{Oseledets}},
  \bibinfo{author}{\bibfnamefont{A.-H.} \bibnamefont{Phan}},
  \bibinfo{author}{\bibfnamefont{Q.}~\bibnamefont{Zhao}}, \bibnamefont{and}
  \bibinfo{author}{\bibfnamefont{D.~P.} \bibnamefont{Mandic}},
  \bibinfo{journal}{Foundations and Trends® in Machine Learning}
  \textbf{\bibinfo{volume}{9}}, \bibinfo{pages}{249} (\bibinfo{year}{2016}),
  ISSN \bibinfo{issn}{1935-8237}.

\bibitem[{\citenamefont{Cichocki et~al.}(2017)\citenamefont{Cichocki, Phan,
  Zhao, Lee, Oseledets, Sugiyama, and Mandic}}]{MAL-067}
\bibinfo{author}{\bibfnamefont{A.}~\bibnamefont{Cichocki}},
  \bibinfo{author}{\bibfnamefont{A.-H.} \bibnamefont{Phan}},
  \bibinfo{author}{\bibfnamefont{Q.}~\bibnamefont{Zhao}},
  \bibinfo{author}{\bibfnamefont{N.}~\bibnamefont{Lee}},
  \bibinfo{author}{\bibfnamefont{I.}~\bibnamefont{Oseledets}},
  \bibinfo{author}{\bibfnamefont{M.}~\bibnamefont{Sugiyama}}, \bibnamefont{and}
  \bibinfo{author}{\bibfnamefont{D.~P.} \bibnamefont{Mandic}},
  \bibinfo{journal}{Foundations and Trends® in Machine Learning}
  \textbf{\bibinfo{volume}{9}}, \bibinfo{pages}{431} (\bibinfo{year}{2017}),
  ISSN \bibinfo{issn}{1935-8237}.

\bibitem[{\citenamefont{Glasser et~al.}(2018)\citenamefont{Glasser, Pancotti,
  and Cirac}}]{glasser2018supervised}
\bibinfo{author}{\bibfnamefont{I.}~\bibnamefont{Glasser}},
  \bibinfo{author}{\bibfnamefont{N.}~\bibnamefont{Pancotti}}, \bibnamefont{and}
  \bibinfo{author}{\bibfnamefont{J.~I.} \bibnamefont{Cirac}},
  \bibinfo{journal}{arXiv:1806.05964}  (\bibinfo{year}{2018}).

\bibitem[{\citenamefont{Verstraete et~al.}(2008)\citenamefont{Verstraete, Murg,
  and Cirac}}]{verstraete2008matrix}
\bibinfo{author}{\bibfnamefont{F.}~\bibnamefont{Verstraete}},
  \bibinfo{author}{\bibfnamefont{V.}~\bibnamefont{Murg}}, \bibnamefont{and}
  \bibinfo{author}{\bibfnamefont{J.~I.} \bibnamefont{Cirac}},
  \bibinfo{journal}{Advances in Physics} \textbf{\bibinfo{volume}{57}},
  \bibinfo{pages}{143} (\bibinfo{year}{2008}).

\bibitem[{\citenamefont{Orús}(2014)}]{ORUS2014117}
\bibinfo{author}{\bibfnamefont{R.}~\bibnamefont{Orús}},
  \bibinfo{journal}{Annals of Physics} \textbf{\bibinfo{volume}{349}},
  \bibinfo{pages}{117 } (\bibinfo{year}{2014}), ISSN \bibinfo{issn}{0003-4916}.

\bibitem[{\citenamefont{Ran et~al.}(2017)\citenamefont{Ran, Tirrito, Peng,
  Chen, Su, and Lewenstein}}]{ran2017review}
\bibinfo{author}{\bibfnamefont{S.-J.} \bibnamefont{Ran}},
  \bibinfo{author}{\bibfnamefont{E.}~\bibnamefont{Tirrito}},
  \bibinfo{author}{\bibfnamefont{C.}~\bibnamefont{Peng}},
  \bibinfo{author}{\bibfnamefont{X.}~\bibnamefont{Chen}},
  \bibinfo{author}{\bibfnamefont{G.}~\bibnamefont{Su}}, \bibnamefont{and}
  \bibinfo{author}{\bibfnamefont{M.}~\bibnamefont{Lewenstein}},
  \bibinfo{journal}{arXiv:1708.09213}  (\bibinfo{year}{2017}).

\bibitem[{\citenamefont{Evenbly and Vidal}(2011)}]{Evenbly2011}
\bibinfo{author}{\bibfnamefont{G.}~\bibnamefont{Evenbly}} \bibnamefont{and}
  \bibinfo{author}{\bibfnamefont{G.}~\bibnamefont{Vidal}},
  \bibinfo{journal}{Journal of Statistical Physics}
  \textbf{\bibinfo{volume}{145}}, \bibinfo{pages}{891} (\bibinfo{year}{2011}),
  ISSN \bibinfo{issn}{1572-9613}.

\bibitem[{\citenamefont{Bridgeman and Chubb}(2017)}]{bridgeman2017hand}
\bibinfo{author}{\bibfnamefont{J.~C.} \bibnamefont{Bridgeman}}
  \bibnamefont{and} \bibinfo{author}{\bibfnamefont{C.~T.} \bibnamefont{Chubb}},
  \bibinfo{journal}{Journal of Physics A: Mathematical and Theoretical}
  \textbf{\bibinfo{volume}{50}}, \bibinfo{pages}{223001}
  (\bibinfo{year}{2017}).

\bibitem[{\citenamefont{Schollwöck}(2011)}]{SCHOLLWOCK201196}
\bibinfo{author}{\bibfnamefont{U.}~\bibnamefont{Schollwöck}},
  \bibinfo{journal}{Annals of Physics} \textbf{\bibinfo{volume}{326}},
  \bibinfo{pages}{96 } (\bibinfo{year}{2011}), ISSN \bibinfo{issn}{0003-4916},
  \bibinfo{note}{january 2011 Special Issue}.

\bibitem[{\citenamefont{Cirac and Verstraete}(2009{\natexlab{a}})}]{Cirac_2009}
\bibinfo{author}{\bibfnamefont{J.~I.} \bibnamefont{Cirac}} \bibnamefont{and}
  \bibinfo{author}{\bibfnamefont{F.}~\bibnamefont{Verstraete}},
  \bibinfo{journal}{Journal of Physics A: Mathematical and Theoretical}
  \textbf{\bibinfo{volume}{42}}, \bibinfo{pages}{504004}
  (\bibinfo{year}{2009}{\natexlab{a}}).

\bibitem[{\citenamefont{Stoudenmire and Schwab}(2016)}]{NIPS2016_6211}
\bibinfo{author}{\bibfnamefont{E.}~\bibnamefont{Stoudenmire}} \bibnamefont{and}
  \bibinfo{author}{\bibfnamefont{D.~J.} \bibnamefont{Schwab}}, in
  \emph{\bibinfo{booktitle}{Advances in Neural Information Processing Systems
  29}}, edited by \bibinfo{editor}{\bibfnamefont{D.~D.} \bibnamefont{Lee}},
  \bibinfo{editor}{\bibfnamefont{M.}~\bibnamefont{Sugiyama}},
  \bibinfo{editor}{\bibfnamefont{U.~V.} \bibnamefont{Luxburg}},
  \bibinfo{editor}{\bibfnamefont{I.}~\bibnamefont{Guyon}}, \bibnamefont{and}
  \bibinfo{editor}{\bibfnamefont{R.}~\bibnamefont{Garnett}}
  (\bibinfo{publisher}{Curran Associates, Inc.}, \bibinfo{year}{2016}), pp.
  \bibinfo{pages}{4799--4807}.

\bibitem[{\citenamefont{Liu et~al.}(2017)\citenamefont{Liu, Ran, Wittek, Peng,
  Garc{\'\i}a, Su, and Lewenstein}}]{liu2017machine}
\bibinfo{author}{\bibfnamefont{D.}~\bibnamefont{Liu}},
  \bibinfo{author}{\bibfnamefont{S.-J.} \bibnamefont{Ran}},
  \bibinfo{author}{\bibfnamefont{P.}~\bibnamefont{Wittek}},
  \bibinfo{author}{\bibfnamefont{C.}~\bibnamefont{Peng}},
  \bibinfo{author}{\bibfnamefont{R.~B.} \bibnamefont{Garc{\'\i}a}},
  \bibinfo{author}{\bibfnamefont{G.}~\bibnamefont{Su}}, \bibnamefont{and}
  \bibinfo{author}{\bibfnamefont{M.}~\bibnamefont{Lewenstein}},
  \bibinfo{journal}{arXiv:1710.04833}  (\bibinfo{year}{2017}).

\bibitem[{\citenamefont{Liu et~al.}(2018)\citenamefont{Liu, Zhang, Lewenstein,
  and Ran}}]{liu2018learning}
\bibinfo{author}{\bibfnamefont{Y.}~\bibnamefont{Liu}},
  \bibinfo{author}{\bibfnamefont{X.}~\bibnamefont{Zhang}},
  \bibinfo{author}{\bibfnamefont{M.}~\bibnamefont{Lewenstein}},
  \bibnamefont{and} \bibinfo{author}{\bibfnamefont{S.-J.} \bibnamefont{Ran}},
  \bibinfo{journal}{arXiv:1803.09111}  (\bibinfo{year}{2018}).

\bibitem[{\citenamefont{Han et~al.}(2018)\citenamefont{Han, Wang, Fan, Wang,
  and Zhang}}]{PhysRevX.8.031012}
\bibinfo{author}{\bibfnamefont{Z.-Y.} \bibnamefont{Han}},
  \bibinfo{author}{\bibfnamefont{J.}~\bibnamefont{Wang}},
  \bibinfo{author}{\bibfnamefont{H.}~\bibnamefont{Fan}},
  \bibinfo{author}{\bibfnamefont{L.}~\bibnamefont{Wang}}, \bibnamefont{and}
  \bibinfo{author}{\bibfnamefont{P.}~\bibnamefont{Zhang}},
  \bibinfo{journal}{Phys. Rev. X} \textbf{\bibinfo{volume}{8}},
  \bibinfo{pages}{031012} (\bibinfo{year}{2018}).

\bibitem[{\citenamefont{Cheng et~al.}(2018)\citenamefont{Cheng, Chen, and
  Wang}}]{e20080583}
\bibinfo{author}{\bibfnamefont{S.}~\bibnamefont{Cheng}},
  \bibinfo{author}{\bibfnamefont{J.}~\bibnamefont{Chen}}, \bibnamefont{and}
  \bibinfo{author}{\bibfnamefont{L.}~\bibnamefont{Wang}},
  \bibinfo{journal}{Entropy} \textbf{\bibinfo{volume}{20}}
  (\bibinfo{year}{2018}), ISSN \bibinfo{issn}{1099-4300}.

\bibitem[{\citenamefont{Pestun and Vlassopoulos}(2017)}]{pestun2017tensor}
\bibinfo{author}{\bibfnamefont{V.}~\bibnamefont{Pestun}} \bibnamefont{and}
  \bibinfo{author}{\bibfnamefont{Y.}~\bibnamefont{Vlassopoulos}},
  \bibinfo{journal}{arXiv:1710.10248}  (\bibinfo{year}{2017}).

\bibitem[{\citenamefont{Cheng et~al.}(2019)\citenamefont{Cheng, Wang, Xiang,
  and Zhang}}]{cheng2019tree}
\bibinfo{author}{\bibfnamefont{S.}~\bibnamefont{Cheng}},
  \bibinfo{author}{\bibfnamefont{L.}~\bibnamefont{Wang}},
  \bibinfo{author}{\bibfnamefont{T.}~\bibnamefont{Xiang}}, \bibnamefont{and}
  \bibinfo{author}{\bibfnamefont{P.}~\bibnamefont{Zhang}},
  \bibinfo{journal}{arXiv:1901.02217}  (\bibinfo{year}{2019}).

\bibitem[{\citenamefont{{Chen} et~al.}(2018)\citenamefont{{Chen}, {Guo}, and
  {Pan}}}]{8406391}
\bibinfo{author}{\bibfnamefont{Y.~W.} \bibnamefont{{Chen}}},
  \bibinfo{author}{\bibfnamefont{K.}~\bibnamefont{{Guo}}}, \bibnamefont{and}
  \bibinfo{author}{\bibfnamefont{Y.}~\bibnamefont{{Pan}}}, in
  \emph{\bibinfo{booktitle}{2018 33rd Youth Academic Annual Conference of
  Chinese Association of Automation (YAC)}} (\bibinfo{year}{2018}), pp.
  \bibinfo{pages}{311--315}.

\bibitem[{\citenamefont{Guo et~al.}(2018)\citenamefont{Guo, Jie, Lu, and
  Poletti}}]{PhysRevE.98.042114}
\bibinfo{author}{\bibfnamefont{C.}~\bibnamefont{Guo}},
  \bibinfo{author}{\bibfnamefont{Z.}~\bibnamefont{Jie}},
  \bibinfo{author}{\bibfnamefont{W.}~\bibnamefont{Lu}}, \bibnamefont{and}
  \bibinfo{author}{\bibfnamefont{D.}~\bibnamefont{Poletti}},
  \bibinfo{journal}{Phys. Rev. E} \textbf{\bibinfo{volume}{98}},
  \bibinfo{pages}{042114} (\bibinfo{year}{2018}).

\bibitem[{\citenamefont{Kak}(1995)}]{kak1995quantum}
\bibinfo{author}{\bibfnamefont{S.}~\bibnamefont{Kak}},
  \bibinfo{journal}{Information Sciences} \textbf{\bibinfo{volume}{83}},
  \bibinfo{pages}{143} (\bibinfo{year}{1995}).

\bibitem[{\citenamefont{Benedetti et~al.}(2018)\citenamefont{Benedetti,
  Garcia-Pintos, Nam, and Perdomo-Ortiz}}]{benedetti2018generative}
\bibinfo{author}{\bibfnamefont{M.}~\bibnamefont{Benedetti}},
  \bibinfo{author}{\bibfnamefont{D.}~\bibnamefont{Garcia-Pintos}},
  \bibinfo{author}{\bibfnamefont{Y.}~\bibnamefont{Nam}}, \bibnamefont{and}
  \bibinfo{author}{\bibfnamefont{A.}~\bibnamefont{Perdomo-Ortiz}},
  \bibinfo{journal}{arXiv:1801.07686}  (\bibinfo{year}{2018}).

\bibitem[{\citenamefont{Mitarai et~al.}(2018)\citenamefont{Mitarai, Negoro,
  Kitagawa, and Fujii}}]{mitarai2018quantum}
\bibinfo{author}{\bibfnamefont{K.}~\bibnamefont{Mitarai}},
  \bibinfo{author}{\bibfnamefont{M.}~\bibnamefont{Negoro}},
  \bibinfo{author}{\bibfnamefont{M.}~\bibnamefont{Kitagawa}}, \bibnamefont{and}
  \bibinfo{author}{\bibfnamefont{K.}~\bibnamefont{Fujii}},
  \bibinfo{journal}{arXiv:1803.00745}  (\bibinfo{year}{2018}).

\bibitem[{\citenamefont{Rebentrost et~al.}(2014)\citenamefont{Rebentrost,
  Mohseni, and Lloyd}}]{PhysRevLett.113.130503}
\bibinfo{author}{\bibfnamefont{P.}~\bibnamefont{Rebentrost}},
  \bibinfo{author}{\bibfnamefont{M.}~\bibnamefont{Mohseni}}, \bibnamefont{and}
  \bibinfo{author}{\bibfnamefont{S.}~\bibnamefont{Lloyd}},
  \bibinfo{journal}{Phys. Rev. Lett.} \textbf{\bibinfo{volume}{113}},
  \bibinfo{pages}{130503} (\bibinfo{year}{2014}).

\bibitem[{\citenamefont{Lloyd et~al.}(2013)\citenamefont{Lloyd, Mohseni, and
  Rebentrost}}]{lloyd2013quantum}
\bibinfo{author}{\bibfnamefont{S.}~\bibnamefont{Lloyd}},
  \bibinfo{author}{\bibfnamefont{M.}~\bibnamefont{Mohseni}}, \bibnamefont{and}
  \bibinfo{author}{\bibfnamefont{P.}~\bibnamefont{Rebentrost}},
  \bibinfo{journal}{arXiv:1307.0411}  (\bibinfo{year}{2013}).

\bibitem[{\citenamefont{Liu and Wang}(2018)}]{PhysRevA.98.062324}
\bibinfo{author}{\bibfnamefont{J.-G.} \bibnamefont{Liu}} \bibnamefont{and}
  \bibinfo{author}{\bibfnamefont{L.}~\bibnamefont{Wang}},
  \bibinfo{journal}{Phys. Rev. A} \textbf{\bibinfo{volume}{98}},
  \bibinfo{pages}{062324} (\bibinfo{year}{2018}).

\bibitem[{\citenamefont{Zeng et~al.}(2018)\citenamefont{Zeng, Wu, Liu, Wang,
  and Hu}}]{zeng2018learning}
\bibinfo{author}{\bibfnamefont{J.}~\bibnamefont{Zeng}},
  \bibinfo{author}{\bibfnamefont{Y.}~\bibnamefont{Wu}},
  \bibinfo{author}{\bibfnamefont{J.-G.} \bibnamefont{Liu}},
  \bibinfo{author}{\bibfnamefont{L.}~\bibnamefont{Wang}}, \bibnamefont{and}
  \bibinfo{author}{\bibfnamefont{J.}~\bibnamefont{Hu}},
  \bibinfo{journal}{arXiv:1808.03425}  (\bibinfo{year}{2018}).

\bibitem[{\citenamefont{McClean et~al.}(2016)\citenamefont{McClean, Romero,
  Babbush, and Aspuru-Guzik}}]{mcclean2016theory}
\bibinfo{author}{\bibfnamefont{J.~R.} \bibnamefont{McClean}},
  \bibinfo{author}{\bibfnamefont{J.}~\bibnamefont{Romero}},
  \bibinfo{author}{\bibfnamefont{R.}~\bibnamefont{Babbush}}, \bibnamefont{and}
  \bibinfo{author}{\bibfnamefont{A.}~\bibnamefont{Aspuru-Guzik}},
  \bibinfo{journal}{New Journal of Physics} \textbf{\bibinfo{volume}{18}},
  \bibinfo{pages}{023023} (\bibinfo{year}{2016}).

\bibitem[{\citenamefont{Farhi et~al.}(2014)\citenamefont{Farhi, Goldstone, and
  Gutmann}}]{farhi2014quantum}
\bibinfo{author}{\bibfnamefont{E.}~\bibnamefont{Farhi}},
  \bibinfo{author}{\bibfnamefont{J.}~\bibnamefont{Goldstone}},
  \bibnamefont{and} \bibinfo{author}{\bibfnamefont{S.}~\bibnamefont{Gutmann}},
  \bibinfo{journal}{arXiv:1411.4028}  (\bibinfo{year}{2014}).

\bibitem[{\citenamefont{Farhi and Neven}(2018)}]{farhi2018classification}
\bibinfo{author}{\bibfnamefont{E.}~\bibnamefont{Farhi}} \bibnamefont{and}
  \bibinfo{author}{\bibfnamefont{H.}~\bibnamefont{Neven}},
  \bibinfo{journal}{arXiv:1802.06002}  (\bibinfo{year}{2018}).

\bibitem[{\citenamefont{Schuld et~al.}(2015)\citenamefont{Schuld, Sinayskiy,
  and Petruccione}}]{schuld2015introduction}
\bibinfo{author}{\bibfnamefont{M.}~\bibnamefont{Schuld}},
  \bibinfo{author}{\bibfnamefont{I.}~\bibnamefont{Sinayskiy}},
  \bibnamefont{and}
  \bibinfo{author}{\bibfnamefont{F.}~\bibnamefont{Petruccione}},
  \bibinfo{journal}{Contemporary Physics} \textbf{\bibinfo{volume}{56}},
  \bibinfo{pages}{172} (\bibinfo{year}{2015}).

\bibitem[{\citenamefont{Arunachalam and
  de~Wolf}(2017)}]{Arunachalam:2017:GCS:3106700.3106710}
\bibinfo{author}{\bibfnamefont{S.}~\bibnamefont{Arunachalam}} \bibnamefont{and}
  \bibinfo{author}{\bibfnamefont{R.}~\bibnamefont{de~Wolf}},
  \bibinfo{journal}{SIGACT News} \textbf{\bibinfo{volume}{48}},
  \bibinfo{pages}{41} (\bibinfo{year}{2017}), ISSN \bibinfo{issn}{0163-5700}.

\bibitem[{\citenamefont{Wiebe et~al.}(2012)\citenamefont{Wiebe, Braun, and
  Lloyd}}]{PhysRevLett.109.050505}
\bibinfo{author}{\bibfnamefont{N.}~\bibnamefont{Wiebe}},
  \bibinfo{author}{\bibfnamefont{D.}~\bibnamefont{Braun}}, \bibnamefont{and}
  \bibinfo{author}{\bibfnamefont{S.}~\bibnamefont{Lloyd}},
  \bibinfo{journal}{Phys. Rev. Lett.} \textbf{\bibinfo{volume}{109}},
  \bibinfo{pages}{050505} (\bibinfo{year}{2012}).

\bibitem[{\citenamefont{Cincio et~al.}(2008)\citenamefont{Cincio, Dziarmaga,
  and Rams}}]{PhysRevLett.100.240603}
\bibinfo{author}{\bibfnamefont{L.}~\bibnamefont{Cincio}},
  \bibinfo{author}{\bibfnamefont{J.}~\bibnamefont{Dziarmaga}},
  \bibnamefont{and} \bibinfo{author}{\bibfnamefont{M.~M.} \bibnamefont{Rams}},
  \bibinfo{journal}{Phys. Rev. Lett.} \textbf{\bibinfo{volume}{100}},
  \bibinfo{pages}{240603} (\bibinfo{year}{2008}).

\bibitem[{\citenamefont{Huggins et~al.}(2019)\citenamefont{Huggins, Patil,
  Mitchell, Whaley, and Stoudenmire}}]{huggins2018towards}
\bibinfo{author}{\bibfnamefont{W.}~\bibnamefont{Huggins}},
  \bibinfo{author}{\bibfnamefont{P.}~\bibnamefont{Patil}},
  \bibinfo{author}{\bibfnamefont{B.}~\bibnamefont{Mitchell}},
  \bibinfo{author}{\bibfnamefont{K.~B.} \bibnamefont{Whaley}},
  \bibnamefont{and} \bibinfo{author}{\bibfnamefont{E.~M.}
  \bibnamefont{Stoudenmire}}, \bibinfo{journal}{Quantum Science and Technology}
  \textbf{\bibinfo{volume}{4}}, \bibinfo{pages}{024001} (\bibinfo{year}{2019}).

\bibitem[{\citenamefont{{Deng}}(2012)}]{6296535}
\bibinfo{author}{\bibfnamefont{L.}~\bibnamefont{{Deng}}},
  \bibinfo{journal}{IEEE Signal Processing Magazine}
  \textbf{\bibinfo{volume}{29}}, \bibinfo{pages}{141} (\bibinfo{year}{2012}),
  ISSN \bibinfo{issn}{1053-5888}.

\bibitem[{\citenamefont{Xiao et~al.}(2017)\citenamefont{Xiao, Rasul, and
  Vollgraf}}]{xiao2017fashion}
\bibinfo{author}{\bibfnamefont{H.}~\bibnamefont{Xiao}},
  \bibinfo{author}{\bibfnamefont{K.}~\bibnamefont{Rasul}}, \bibnamefont{and}
  \bibinfo{author}{\bibfnamefont{R.}~\bibnamefont{Vollgraf}},
  \bibinfo{journal}{arXiv:1708.07747}  (\bibinfo{year}{2017}).

\bibitem[{\citenamefont{Cortes and Vapnik}(1995)}]{Cortes1995}
\bibinfo{author}{\bibfnamefont{C.}~\bibnamefont{Cortes}} \bibnamefont{and}
  \bibinfo{author}{\bibfnamefont{V.}~\bibnamefont{Vapnik}},
  \bibinfo{journal}{Machine Learning} \textbf{\bibinfo{volume}{20}},
  \bibinfo{pages}{273} (\bibinfo{year}{1995}), ISSN \bibinfo{issn}{1573-0565}.

\bibitem[{\citenamefont{Rish et~al.}(2001)}]{rish2001empirical}
\bibinfo{author}{\bibfnamefont{I.}~\bibnamefont{Rish}} \bibnamefont{et~al.}, in
  \emph{\bibinfo{booktitle}{IJCAI 2001 Workshop on Empirical Methods in
  Artificial Intelligence}} (\bibinfo{year}{2001}), vol.~\bibinfo{volume}{3},
  pp. \bibinfo{pages}{41--46}.

\bibitem[{\citenamefont{Novikov et~al.}(2016)\citenamefont{Novikov, Trofimov,
  and Oseledets}}]{novikov2016exponential}
\bibinfo{author}{\bibfnamefont{A.}~\bibnamefont{Novikov}},
  \bibinfo{author}{\bibfnamefont{M.}~\bibnamefont{Trofimov}}, \bibnamefont{and}
  \bibinfo{author}{\bibfnamefont{I.}~\bibnamefont{Oseledets}},
  \bibinfo{journal}{arXiv:1605.03795}  (\bibinfo{year}{2016}).

\bibitem[{\citenamefont{Cirac and
  Verstraete}(2009{\natexlab{b}})}]{cirac2009renormalization}
\bibinfo{author}{\bibfnamefont{J.~I.} \bibnamefont{Cirac}} \bibnamefont{and}
  \bibinfo{author}{\bibfnamefont{F.}~\bibnamefont{Verstraete}},
  \bibinfo{journal}{Journal of Physics A: Mathematical and Theoretical}
  \textbf{\bibinfo{volume}{42}}, \bibinfo{pages}{504004}
  (\bibinfo{year}{2009}{\natexlab{b}}).

\bibitem[{\citenamefont{Kullback and Leibler}(1951)}]{10.2307/2236703}
\bibinfo{author}{\bibfnamefont{S.}~\bibnamefont{Kullback}} \bibnamefont{and}
  \bibinfo{author}{\bibfnamefont{R.~A.} \bibnamefont{Leibler}},
  \bibinfo{journal}{The Annals of Mathematical Statistics}
  \textbf{\bibinfo{volume}{22}}, \bibinfo{pages}{79} (\bibinfo{year}{1951}),
  ISSN \bibinfo{issn}{00034851}.

\bibitem[{\citenamefont{Ciresan et~al.}(2012)\citenamefont{Ciresan, Meier, and
  Schmidhuber}}]{Ciresan2012MulticolumnDN}
\bibinfo{author}{\bibfnamefont{D.~C.} \bibnamefont{Ciresan}},
  \bibinfo{author}{\bibfnamefont{U.}~\bibnamefont{Meier}}, \bibnamefont{and}
  \bibinfo{author}{\bibfnamefont{J.}~\bibnamefont{Schmidhuber}},
  \bibinfo{journal}{2012 IEEE Conference on Computer Vision and Pattern
  Recognition} pp. \bibinfo{pages}{3642--3649} (\bibinfo{year}{2012}).

\bibitem[{\citenamefont{{Bhatnagar} et~al.}(2017)\citenamefont{{Bhatnagar},
  {Ghosal}, and {Kolekar}}}]{8313740}
\bibinfo{author}{\bibfnamefont{S.}~\bibnamefont{{Bhatnagar}}},
  \bibinfo{author}{\bibfnamefont{D.}~\bibnamefont{{Ghosal}}}, \bibnamefont{and}
  \bibinfo{author}{\bibfnamefont{M.~H.} \bibnamefont{{Kolekar}}}, in
  \emph{\bibinfo{booktitle}{2017 Fourth International Conference on Image
  Information Processing (ICIIP)}} (\bibinfo{year}{2017}), pp.
  \bibinfo{pages}{1--6}.

\bibitem[{\citenamefont{Mercer and Forsyth}(1909)}]{doi:10.1098/rsta.1909.0016}
\bibinfo{author}{\bibfnamefont{J.}~\bibnamefont{Mercer}} \bibnamefont{and}
  \bibinfo{author}{\bibfnamefont{A.~R.} \bibnamefont{Forsyth}},
  \bibinfo{journal}{Philosophical Transactions of the Royal Society of London.
  Series A, Containing Papers of a Mathematical or Physical Character}
  \textbf{\bibinfo{volume}{209}}, \bibinfo{pages}{415} (\bibinfo{year}{1909}).

\bibitem[{\citenamefont{Leslie et~al.}(2001)\citenamefont{Leslie, Eskin, and
  Noble}}]{leslie2001spectrum}
\bibinfo{author}{\bibfnamefont{C.}~\bibnamefont{Leslie}},
  \bibinfo{author}{\bibfnamefont{E.}~\bibnamefont{Eskin}}, \bibnamefont{and}
  \bibinfo{author}{\bibfnamefont{W.~S.} \bibnamefont{Noble}}, in
  \emph{\bibinfo{booktitle}{Biocomputing 2002}} (\bibinfo{publisher}{World
  Scientific}, \bibinfo{year}{2001}), pp. \bibinfo{pages}{564--575}.

\bibitem[{\citenamefont{R{\'e}nyi et~al.}(1961)}]{renyi1961measures}
\bibinfo{author}{\bibfnamefont{A.}~\bibnamefont{R{\'e}nyi}}
  \bibnamefont{et~al.}, in \emph{\bibinfo{booktitle}{Proceedings of the Fourth
  Berkeley Symposium on Mathematical Statistics and Probability, Volume 1:
  Contributions to the Theory of Statistics}} (\bibinfo{organization}{The
  Regents of the University of California}, \bibinfo{year}{1961}).

\bibitem[{\citenamefont{Or\'us and Vidal}(2008)}]{PhysRevB.78.155117}
\bibinfo{author}{\bibfnamefont{R.}~\bibnamefont{Or\'us}} \bibnamefont{and}
  \bibinfo{author}{\bibfnamefont{G.}~\bibnamefont{Vidal}},
  \bibinfo{journal}{Phys. Rev. B} \textbf{\bibinfo{volume}{78}},
  \bibinfo{pages}{155117} (\bibinfo{year}{2008}).

\bibitem[{\citenamefont{Perez-Garcia et~al.}(2006)\citenamefont{Perez-Garcia,
  Verstraete, Wolf, and Cirac}}]{perez2006matrix}
\bibinfo{author}{\bibfnamefont{D.}~\bibnamefont{Perez-Garcia}},
  \bibinfo{author}{\bibfnamefont{F.}~\bibnamefont{Verstraete}},
  \bibinfo{author}{\bibfnamefont{M.~M.} \bibnamefont{Wolf}}, \bibnamefont{and}
  \bibinfo{author}{\bibfnamefont{J.~I.} \bibnamefont{Cirac}},
  \bibinfo{journal}{arXiv preprint quant-ph/0608197}  (\bibinfo{year}{2006}).

\bibitem[{\citenamefont{Tagliacozzo et~al.}(2008)\citenamefont{Tagliacozzo,
  de~Oliveira, Iblisdir, and Latorre}}]{PhysRevB.78.024410}
\bibinfo{author}{\bibfnamefont{L.}~\bibnamefont{Tagliacozzo}},
  \bibinfo{author}{\bibfnamefont{T.~R.} \bibnamefont{de~Oliveira}},
  \bibinfo{author}{\bibfnamefont{S.}~\bibnamefont{Iblisdir}}, \bibnamefont{and}
  \bibinfo{author}{\bibfnamefont{J.~I.} \bibnamefont{Latorre}},
  \bibinfo{journal}{Phys. Rev. B} \textbf{\bibinfo{volume}{78}},
  \bibinfo{pages}{024410} (\bibinfo{year}{2008}).

\end{thebibliography}
    	\end{document}